# Comparison of Fuzzy and Neuro Fuzzy Image Fusion Techniques and its Applications

Srinivasa Rao D
Dept. of IT
VNRVJIET
Hyderabad

Seetha M
Dept. of CSE
GNITS
Hyderabad

Krishna Prasad MHM
Dept. of IT
JNTU Vizianagaram campus
Vizianagaram

## ABSTRACT
Image fusion is the process of integrating multiple images of the same scene into a single fused image to reduce uncertainty and minimizing redundancy while extracting all the useful information from the source images. Image fusion process is required for different applications like medical imaging, remote sensing, medical imaging, machine vision, biometrics and military applications where quality and critical information is required. In this paper, image fusion using fuzzy and neuro fuzzy logic approaches utilized to fuse images from different sensors, in order to enhance visualization. The proposed work further explores comparison between fuzzy based image fusion and neuro fuzzy fusion technique along with quality evaluation indices for image fusion like image quality index, mutual information measure, fusion factor, fusion symmetry, fusion index, root mean square error, peak signal to noise ratio, entropy, correlation coefficient and spatial frequency. Experimental results obtained from fusion process prove that the use of the neuro fuzzy based image fusion approach shows better performance in first two test cases while in the third test case fuzzy based image fusion technique gives better results.

## 1. INTRODUCTION
Image fusion is the process to combine information from two or more images of a scene into a single composite image that is more informative and is more suitable for visual perception or computer processing. It has been used in great many disciplines like medical imaging, remote sensing, navigation aid, machine vision, automatic change detection, biometrics and military applications etc. Multisensor image fusion for surveillance systems is proposed in which fuzzy logic approach utilized to fuse images from different sensors, in order to enhance visualization for surveillance [1]. In [2] urban remote image fusion using fuzzy rules approach utilized to refine the resolution of urban multi-spectral images using the corresponding high-resolution panchromatic images. After the decomposition of two input images by wavelet transform three texture features are extracted and then a fuzzy fusion rule is used to merge wavelet coefficients from the two images according to the extracted features. In [3] image fusion algorithm based on fuzzy logic and wavelet, aimed at the visible and infrared image fusion and address an algorithm based on the discrete wavelet transform and fuzzy logic. In [3] the technique created two fuzzy relations, and estimated the importance of every wavelet coefficient with fuzzy reasoning. In [4] an Iterative Fuzzy and Neuro Fuzzy approach proposed for fusing medical images and remote sensing images and found that the technique very useful in medical imaging and other areas, where quality of image is more important than the real time application. In [ 5] a new method is proposed for Pixel-Level Multisensor image fusion based on Fuzzy Logic in which the membership function and fuzzy rules of the new algorithm is defined using the Fuzzy Inference System. A fuzzy radial basis function neural networks is used to perform auto-adaptive image fusion and in experiment multimodal medical image fusion based on gradient pyramid is performed for comparison [6]. In [7] a novel method is proposed using combine framework of wavelet transform and fuzzy logic and it provides novel tradeoff solution between the spectral and spatial fidelity and preserves more detail spectral and spatial information. Pixel & Feature Level Multi-Resolution Image Fusion based on Fuzzy logic in which images are first segmented into regions with fuzzy clustering and are then fed into a fusion system, based on fuzzy if-then rules [8]. In [9] a fusion algorithm based on neuro-fuzzy logic is presented, and utilized hybrid algorithm which mixes back propagation algorithm with least mean square (LMS) algorithm to train the parameters of membership function. In [10] a image fusion using fuzzy logic proposed for applications like automotive, medical and other areas. In [11] a fuzzy logic approach proposed to fuse images from different sensors, in order to enhance visualization and work further explores the comparison between image fusion using wavelet transform and fuzzy logic approach

## 2. FUZZY APPROACH ELEMENTS
### 2.1 Fuzzy Logic
The importance of fuzzy logic derives from the fact that most modes of human thinking and especially common sense reasoning are approximate in nature. The essential features of fuzzy logic as founded by Zader Lotfi are as follows

- In fuzzy logic everything is a matter of degree
- Any logical system can be fuzzified
- In fuzzy logic, knowledge is interpreted as a collection of elastic or, equivalently, fuzzy constraint on a collection of variables
- Inference is viewed as a process of propagation of elastic constraints

The goal of fuzzy approach used to describe powerful characteristics of fuzzy sets when used specially for image information processing. It has been chosen to focus on the following points:

- Fuzzy sets are to represent spatial information in images along with its imprecision
- Operations recently generalized to fuzzy sets in order to manage spatial information and
- Information fusion using fuzzy combination operators
- Fast computation using fuzzy number operations.

In [12] Praveena proposed a method that applications using fuzzy sets give promising results from low level processing to higher-level image interpretation. Applications of fuzzy logic and fuzzy sets to image processing are quite recent, compared with other functions like control, but they give rise now to a large development through ability of fuzzy sets to represent and to manage imprecise spatial information. However, fuzzy sets do have a high interest for image processing, which deserves a deeper attention.

## 2.2 Fuzzy Sets
In [13] Zadeh proposed that fuzzy set is a class of objects with a continuum of grades of membership. Fuzzy set is characterized by a membership function which assigns to each object a grade of membership ranging between zero and one. It was introduced as a mean to model the vagueness and ambiguity in complex systems. The idea of fuzzy sets is simple and natural.

## 2.3 Membership Functions
The membership function is a graphical representation of the magnitude of participation of each input in the input space. Input space is often referred as the universe of discourse or universal set, which contain all the possible elements of concern in each particular application. It associates a weighting with each of the inputs that are processed, define functional overlap between inputs, and ultimately determines an output response. The rules use the input membership values as weighting factors to determine their influence on the fuzzy output sets of the final output conclusion. Once the functions are inferred, scaled, and combined, they are defuzzified into a crisp output, which drives the system. There are different memberships functions associated with each input and output response [9].

## 2.4 Fuzzy Rules
Human beings make decisions based on rules. Fuzzy machines, which always tend to mimic the behavior of man, work the same way. However, the decision and the means of choosing that decision are replaced by fuzzy sets and the rules are replaced by fuzzy rules. Fuzzy rules also operate using a series of if-then statements. For instance, if X then A, if Y then B, where A and B are all sets of X and Y. Fuzzy rules define fuzzy patches, which is the key idea in fuzzy logic.

## 3. FUZZY BASED IMAGE FUSION
Fuzzy based image fusion requires that some basic components to be discussed

### 3.1 Fuzzy Logic in Image Processing
Fuzzy image processing is not a unique theory. Fuzzy image processing is the collection of all approaches that understand, represent and process the images, their segments and features as fuzzy sets. The representation and processing depend on the selected fuzzy technique and on the problem to be solved. It has three main stages:

- Image fuzzification((Using membership functions to graphically describe a situation)
- Modification of membership values((Application of fuzzy rules)
- Image defuzzification((Obtaining the crisp or actual results)

The coding of image data (fuzzification) and decoding of the results (defuzzification) are steps that make possible to process images with fuzzy techniques. The main power of fuzzy image processing is in the middle step (modification of membership values). After the image data are transformed from gray-level plane to the membership plane (fuzzification), appropriate fuzzy techniques modify the membership values. This can be a fuzzy clustering, a fuzzy rule-based approach, a fuzzy integration approach and so on [14].

### 3.2 Steps in Fuzzy Image Fusion
The original image in the gray level plane is subjected to fuzzification and the modification of membership functions is carried out in the membership plane. The result is the output image obtained after the defuzzification process.

The algorithm for pixel-level image fusion using fuzzy logic is given as follows [15].
  a) Read first image in variable I1 and find its size (rows: r1, columns: c1).
  b) Read second image in variable I2 and find its size (rows: r2, columns: c2).
  c) Variables I1 and I2 are images in matrix form where each pixel gray level value is in the range from 0 to 255.
  d) Compare rows and columns of both input images. If these two images are not of the same size, select the portion, which are of same size.
  e) Convert the images in column form which has C = r1×c1 entries.
  f) Make a fuzzy inference system file, which has two input images.
  g) Decide number and type of membership functions for both the input images by tuning the membership functions.
  h) Input images in antecedent are resolved to a degree of membership ranging 0 to 255.
  i) Make fuzzy if-then rules for input images, which resolve those two antecedents to a single number from 0 to 255.
  j) For num = 1 to C in steps of 1, apply fuzzification using the rules developed above on the corresponding pixel gray level values of the input images, which gives fuzzy sets represented by

      membership functions and results in output image in column format.
  k) Convert the column form to matrix form and display the fused image.

Membership functions and rules used in the fuzzy system

1. if (input1 is mf1) and (input2 is mf2) then (output1 is mf1)
2. if (input1 is mf2) and (input2 is mf2 then (output1 is mf2)
3. if (input1 is mf2) and (input2 is mf2) then (output1 is mf2)
4. if (input1 is mf3) or (input2 is mf2) then (output1 is mf3)
5. if (input1 is mf1) and (input2 is mf3) then (output1 is mf1)
6. if (input1 is mf3) or (input2 is mf3) then (output1 is mf2)

## 4. NEURO FUZZY BASED IMAGE FUSION

Neuro fuzzy logic based image fusion requires some fundamentals to be discussed.

### 4.1 Neural Network

Neural Network (NN), is a natural propensity for storing experiential knowledge and making it available for use. NNs can provide suitable solutions for problems, which are generally characterized by non-linearity, high dimensionality noisy, complex, imprecise, imperfect or error prone sensor data, and lack of a clearly stated mathematical solution or algorithm. A key benefit of NN is that a model of the system or subject can be built just from the data.

### 4.2 Network Properties

The topology of NN refers to its framework as well as its inter-connection scheme. The framework is often specified by the number of layers and the number of nodes per layer.

### 4.3 Neuro Fuzzy Logic

In the field of artificial intelligence, Neuro-Fuzzy refers to combinations of artificial neural networks and fuzzy logic. Neuro-Fuzzy composite results in a hybrid intelligent system that synergizes these two techniques by combining the human-like reasoning style of fuzzy systems with the learning and connectionist structure of neural networks. Neuro-Fuzzy hybridization is widely termed as Fuzzy Neural Network (FNN) or Neuro-Fuzzy System (NFS) in the literature. Neuro-Fuzzy system incorporates the human-like reasoning style of fuzzy systems through the use of fuzzy sets and a linguistic model consisting of a set of IF-THEN fuzzy rules.

The strength of neuro-fuzzy systems involves two contradictory requirements in fuzzy modeling: interpretability versus accuracy. In practice, one of the two properties prevails. The Neuro-Fuzzy in fuzzy modeling research field is divided into two areas: linguistic fuzzy modeling that is focused on interpretability, mainly the Mamdani model; and precise fuzzy modeling that is focused on accuracy, mainly the Takagi-Sugeno-Kang (TSK) model.

A neuro-fuzzy system is based on a fuzzy system which is trained by a learning algorithm derived from neural network theory. The learning procedure operates on local information, and causes only local modifications in the underlying fuzzy system.

A Neuro-Fuzzy system can be always (i.e.\ before, during and after learning) interpreted as a system of fuzzy rules. It is also possible to create the system out of training data from scratch, as it is possible to initialize it by prior knowledge in form of fuzzy rules.

The learning procedure of a Neuro-Fuzzy system takes the semantical properties of the underlying fuzzy system into account. This results in constraints on the possible modifications applicable to the system parameters.

Neural networks are used to induce knowledge or functional relationships from instances of sampled data. The system can be trained from the input data. This approach utilises 'back propagation' algorithm to train the parameters of membership function. The fused images based on neuro-fuzzy logic not only reserves more texture features, but also enhances the information characteristics of two original images. Once we get a fused image, it is further fused with the one or the other input images to get a better quality image [16]

### 4.4 Advantages of Neuro Fuzzy

Neuro fuzzy logic approach benefits as follows

- Handle any kind of information (numeric, linguistic, logical, etc.)
- Manage imprecise, partial, vague or imperfect information.
- Resolve conflicts by collaboration and aggregation.
- Self-learning, self-organizing and self-tuning capabilities.
- No need of prior knowledge of relationships of data.
- Mimic human decision making process.
- Fast computation using fuzzy number operations.

### 4.5 Algorithm for Neuro Fuzzy Based Image Fusion

The algorithm for pixel-level image fusion using neuro fuzzy logic is given as follows.

a) Read first image in variable I1 and find its size (rows:zl, columns: sl).
b) Read second image in variable I2 and find its size (rows:z2. columns: s2).
c) Variables I1 and I2 are images in matrix form where each pixel value is in the range from 0-255. Use Gray Colormap.
d) Compare rows and columns of both input images. If the two images are not of the same size, select the portion. Which are of same size.
e) Convert the images in column form which has C= zl*sl entries.
f) Form a training data, which is a matrix with three columns and entries in each column are form *0* to 255 insteps of 1.
g) Form a check data. Which is a matrix of Pixels of two input images in column format.

h) Decide number and type of membership functions for both the input images by tuning the membership functions.
i) For training FIS structure is used, which is generated by genfisl command with training data, number of membership functions and type of membership functions as input.
j) To start training, anfis command is used which inputs generated FIS structure and training data and returns trained data.
k) For num=1 to C in steps of one, apply fuzzification using the generated FIS structure with Check data and trained data as inputs which returns output image in column format.
l) Convert the column form to matrix form and display the fused image

# 5. EVALUATION INDICES FOR IMAGE FUSION

Evaluation measures are used to evaluate the quality of the fused image. The fused images are evaluated, taking the following parameters into consideration

## 5.1 Image Quality Index

Image quality index (IQI) measures the similarity between two images (I1 & I2) and its value ranges from -1 to 1. IQI is equal to 1 if both images are identical. IQI measure is given by [17]

$$IQI = \frac{m_{ab} 2xy 2 m_a m_b}{m_a m_b x^2 + y^2 m_a^2 + m_b^2} \quad (1)$$

Where x and y denote the mean values of images I1 and I2 and $m_a^2$, $m_b^2$ and $m_{ab}$ denotes the variance of I1, I2 and covariance of I1 and I2.

## 5.2 Mutual Information Measure

Mutual information measure (MIM) furnishes the amount of information of one image in another. This gives the guidelines for selecting the best fusion method. Given two images *M (i, j) and N (i, j)* and MIM between them is defined as:

$$I_{MN} = \sum_{x,y} P_{MN}(x,y) \log \frac{P_{MN}(x,y)}{P_M(x) P_N(y)} \quad (2)$$

Where, $P_M$ (x) and $P_N$ (y) are the probability density functions in the individual images, and $P_{MN}$ *(x, y)* is joint probability density function.

## 5.3 Fusion Factor

Given two images A and B, and their fused image F, the Fusion factor (FF) is illustrated as [18]

$$FF = I_{AF} + I_{BF} \quad (3)$$

Where $I_{AF}$ and $I_{BF}$ are the MIM values between input images and fused image. A higher value of FF indicates that fused image contains moderately good amount of information present in both the images. However, a high value of FF does not imply that the information from both images is symmetrically fused.

## 5.4 Fusion Symmetry

Fusion symmetry (FS) is an indication of the degree of symmetry in the information content from both the images.

$$FS = abs\left(\frac{I_{AF}}{I_{AF} + I_{BF}} - 0.5\right) \quad (4)$$

The quality of fusion technique depends on the degree of Fusion symmetry. Since FS is the symmetry factor, when the sensors are of good quality, FS should be as low as possible so that the fused image derives features from both input images. If any of the sensors is of low quality then it is better to maximize FS than minimizing it.

## 5.5 Fusion Index

This study proposes a parameter called Fusion index from the factors Fusion symmetry and Fusion factor. The fusion index (FI) is defined as

$$FI = I_{AF} / I_{BF} \quad (5)$$

Where $I_{AF}$ is the mutual information index between multispectral image and fused image and $I_{BF}$ is the mutual information index between panchromatic image and fused image. The quality of fusion technique depends on the degree of fusion index.

## 5.6 Root Mean Square Error

The root mean square error (RMSE) measures the amount of change per pixel due to the processing. The RMSE between a reference image R and the fused image F is given by

$$RMSE = \sqrt{\frac{1}{MN} \sum_{i=1}^{M} \sum_{j=1}^{N} (R(i,j) - F(i,j))} \quad (6)$$

## 5.7 Peak Signal to Noise Ratio

Peak signal to noise ratio (PSNR) can be calculated by using the formula

$$PSNR = 20 \log_{10} \left[\frac{L^2}{MSE}\right] \quad (7)$$

Where MSE is the mean square error and L is the number of gray levels in the image.

## 5.8 Entropy

The entropy of an image is a measure of information content. It is the average number of bits needed to quantize the intensities in the image. It is defined as:

$$E = -\sum (p * \log_2(p)) \quad (8)$$

Where p contains the histogram counts returned from imhist.

## 5.9 Correlation Coefficient

The Correlation Coefficient (CC) method is used to determine how closely the input and output images co-vary. Correlation coefficient is widely used for comparing images. It is widely used in statistical analysis, pattern recognition, and image processing [19].

$$CC = \frac{\sum_{i=1}^{n}(Xi - \overline{X})(Yi - \overline{Y})}{\sqrt{\sum_{i=1}^{n}(Xi - \overline{X})^2}\sqrt{\sum_{i=1}^{n}(Yi - \overline{Y})^2}} \quad (9)$$

Where, $X_i$ is the intensity of the $i^{th}$ pixel in image1, $Y_i$ is the intensity of the $i^{th}$ pixel in image2, X is the mean intensity of image1 and Y is the mean intensity of image2.

## 5.10 Spatial Frequency

The Spatial Frequency (SF) is the number of cycles that fall within one degree of visual angle. A grating of high spatial frequency-- many cycles within each degree of visual angle -- contains narrow bars. A grating of low spatial frequency -- few cycles within each degree of visual angle -- contains wide bars. Because spatial frequency is defined in terms of visual angle, a grating's spatial frequency changes with viewing distance. As this distance decreases, each bar casts a larger image; as a result, the grating's spatial frequency decreases as the distance decreases.

$$SF = \sqrt{(RF)^2 + (CF)^2} \quad (10)$$

This frequency in spatial domain indicates the overall activity level in the fused image [20]

## 6. RESULTS AND DISCUSSIONS

There are many typical applications for image fusion. Modern spectral scanners gather up to several hundred of spectral bands which can be both visualized and processed individually, or which can be fused into a single image, depending on the image analysis task. In this section, input images are fused using fuzzy logic approach. Example 1, Panchromatic and Multispectral images of the Hyderabad city, AP, INDIA are acquired from the IRS 1D LISS III sensor at 05:40:44, Example 2 and Example 3 images are acquired from http://imagefusion.org [21].

The proposed fuzzy and neuro fuzzy based image fusion approaches are implemented in Matlab 10.0. These methods can be scalable and expandable for great many situations like remote sensing, medical imaging and video surveillance etc.. in which membership functions and rules have to define precisely used in fuzzy inference system. In order to evaluate the fusion results obtained from different methods and compare the methods, the assessment measures are employed. The value of each quality assessment parameters of all mentioned fusion approaches are depicted in Table 1.Our experimental results show that neuro fuzzy logic based image fusion approach provides better performance when compared to fuzzy based image fusion for first two examples. Image quality index (IQI), the similarity between reference and fused image (0.9999, 0.9829, and 0.3182) are higher for first two cases when compared to values obtained from fuzzy based fusion technique (0.9758, 0.9824, and 0.8871). The higher values for fusion factor (FF) from first two examples(2.8115, 3.3438, 1.0109) obtained from the neuro fuzzy based fusion approach indicates that fused image contains moderately good amount of information present in both the images compared to FF values (1.0965,2.1329,1.9864) obtained from fuzzy based fusion approach. The amount of information of one image in another, mutual information measure (MIM) values (1.4656,1.5079,0.7634) are also significantly better which shows that neuro fuzzy based fusion method preserves more information compared to fuzzy based image fusion. The other evaluation measures like root mean square error (RMSE) with lower and peak signal to noise ratio (PSNR), Correlation Coefficient (CC) with higher values (0.9459, 0.8979, 0.1265) obtained form neuro fuzzy based fusion approach are also comparatively better for first two cases. The entropy, the amount of information that can be used to characterize the input image (7.2757, 7.3202, 4.4894) are better for two examples obtained from neuro fuzzy based image fusion technique. Through higher values for spatial frequency (25.5698,37.1169,16.9926 ) neuro fuzzy based image fusion preserves more texture features compared to fuzzy based image fusion (10.4567, 16.9749, 25.4711) and also improved the information of two input images. So it is concluded that results obtained from the implementation of neuro fuzzy logic based image fusion approach performs better for first two test cases and fuzzy based image fusion shows better performance for third test case. So further investigation is needed to resolve this issue.

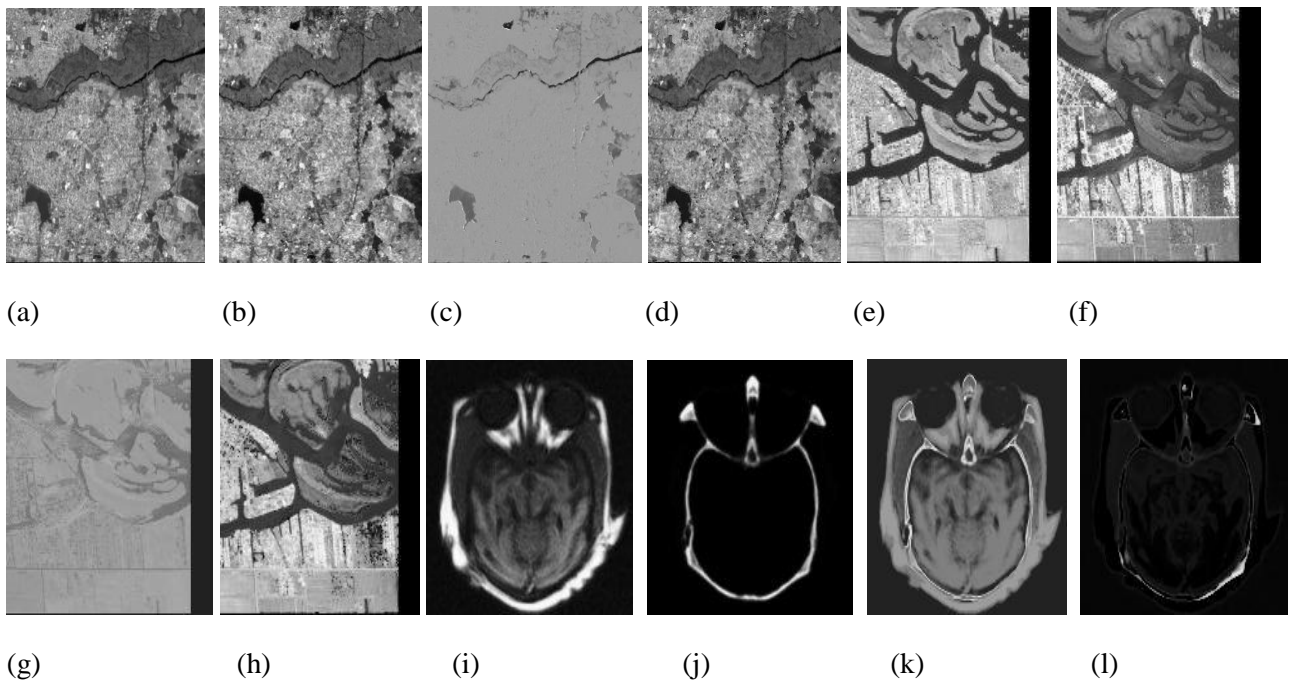

(a)　　　　(b)　　　　(c)　　　　(d)　　　　(e)　　　　(f)

(g)　　　　(h)　　　　(i)　　　　(j)　　　　(k)　　　　(l)

**Fig 1: Some example images (a), (b), (e), (f), (i) and (j): original input images; (c), (g) and (k): fused images by fuzzy logic , (d), (h) and (l): fused images by neuro fuzzy logic respectively**

**Table 1. Evaluation indices for image fusion based on fuzzy and neuro fuzzy logic approaches**

| Method | IQI | MIM | FF | FS | FI | RMSE | PSNR | Entropy | CC | SF |
|---|---|---|---|---|---|---|---|---|---|---|
| Fuzzy Fusion | | | | | | | | | | |
| (Ex 1) | 0.9758 | 0.4628 | 1.0965 | 0.0779 | 0.7303 | 44.8448 | 15.0966 | 4.4881 | 0.5833 | 10.4567 |
| (Ex 2) | 0.9824 | 0.9582 | 2.1379 | 0.0518 | 0.8122 | 48.9935 | 14.3280 | 5.3981 | 0.7598 | 16.9749 |
| (Ex 3) | 0.8871 | 1.5576 | 1.9864 | 0.2841 | 3.6325 | 42.4850 | 15.5661 | 5.8205 | 0.7761 | 25.4711 |
| Neuro Fuzzy Fusion | | | | | | | | | | |
| (Ex 1) | 0.9999 | 1.4656 | 2.8115 | 0.0213 | 1.0837 | 14.9554 | 24.6348 | 7.2757 | 0.9459 | 25.5698 |
| (Ex 2) | 0.9829 | 1.5079 | 3.3438 | 0.0490 | 0.8213 | 34.4662 | 17.3829 | 7.3202 | 0.8979 | 37.1169 |
| (Ex 3) | 0.3182 | 0.7634 | 1.0109 | 0.2431 | 2.8926 | 68.5319 | 11.4129 | 4.4894 | 0.1265 | 16.9926 |

## 7. ACKNOWLEDGMENTS

This work was partially supported by the All India Council for Technical Education, New Delhi, India under Research Promotion Scheme, Grant No. 8023/RID/RPS-80/2010-11.

## 8. CONCLUSIONS

There are a large number of applications like medical imaging, video surveillance and remote sensing etc. that require images with both spatial and spectral resolution as well. In this paper, the potentials of image fusion using fuzzy and neuro fuzzy approaches has been explored along with quality assessment evaluation measures. Fused images are primarily used to human observers for viewing or interpretation and to be further processed by a computer using different image processing techniques**.** All the results obtained and discussed by this method are same scene. The experimental results clearly show that the proposed image fusion using fuzzy logic gives a considerable improvement on the quality of the fusion system and neuro fuzzy based image fusion preserves more texture information. It is hoped that the technique can be further extended to video image processing and for fusion of multiple sensor images and to integrate valid evaluation measures of image fusion. Future work also includes the iterative fuzzy logic and iterative neuro fuzzy logic, which efficiently gives good results.

## 9. REFERENCES

[1] Yi, Z., Ping, Z. 2010 Multisensor Image Fusion Using Fuzzy Logic for Surveillance Systems. IEEE Seventh International Conference on Fuzzy Systems and Discovery, pp. 588-592, shanghai (2010)

[2] Yang, X.H., Huang, F.Z., Liu,G. 2009. Urban Remote Image Fusion Using Fuzzy Rules. IEEE Proceedings of the Eighth International Conference on Machine Learning and Cybernetics, pp. 101-109, (2009)

[3] Mengyu, Z., Yuliang, Y. 2008 . A New image Fusion Algorithm Based on Fuzzy Logic. IEEE International Conference on Intelligent Computation Technology and Automation, pp. 83-86. Changsha (2008)

[4] Ranjan, R., Singh, H., Meitzler, T., Gerhart, G.R. Iterative Image Fusion technique using Fuzzy and Neuro fuzzy Logic and Applications. IEEE Fuzzy Information Processing Society, pp. 706-710, Detroit, USA (2005)

[5] Zhao, L., Xu, B., Tang, W., Chen, Z. A Pixel-Level Multisensor Image Fusion Algorithm based on Fuzzy Logic. LNCS, vol. 3613, pp. 717-720. Springer, Heidelberg (2005)

[6] Wang, Y.P., Dang, J.W., Li, Q., Li, S. Multimodal Medical Image fusion using Fuzzy Radial Basis function Neural Networks, IEEE International Conference on Wavelet Analysis and Pattern Recognition, pp. 778-782. Beijing (2007)

[7]Tanish, Z., Ishit, M., Mukesh, Z. Novel hybrid Multispectral Image Fusion Method using Fuzzy Logic.I.J. Computer Information Systems and Industrial Management Applications. 096-103 (2010)

[8] Bushra, N.K., Anwar, M.M., Haroon, I. Pixel & Feature Level Multi-Resolution Image Fusion based on Fuzzy Logic. ACM Proc. of the 6$^{th}$ WSEAS International Conference on Wavelet analysis & Multirate Systems, pp. 88-91. Romania (2006)

[9]Jionghua,Teng., Suhuan,Wang.,Jingzhou,Zhang., Xue, Wang. Neuro-fuzzy logic based fusion algorithm of medical images. Image and Signal Processing (CISP), pp. 1552 – 1556, 2010

[10]Harpreet, Singh., Jyoti, Raj., Gulsheen, Kaur., Thomas ,Meitzler., Image Fusion using Fuzzy Logic and Applications, IEEE Proceedings International Conference on Fuzzy Systems, 2004

[11]S.R.Dammavalam, S.Maddala, M. H. M. Krishna Prasad., Quality Evaluation Measures of Pixel - Level Image Fusion Using Fuzzy Logic. LNCS 7076,pp. 485-493,2011

[12] Praveena, S.M. Multiresolution Optimization of Image Fusion. National Conference on Recent Trends in Communication and Signal Processing, pp. 111-118. Coimbatore (2009)

[ 13] Zadeh, L.A.. Fuzzy Sets. J. Information and Control. 8, 338-353 (1965

[14] Maruthi, R., Sankarasubramanian, K. Pixel Level Multifocus Image Fusion Based on Fuzzy Logic Approach. J. Information Technology. 7(4), 168-171 (2008)

[15]S.R.Dammavalam, S.Maddala,MHM.KrishnaPrasad. Quality Assessment of Pixel-Level Image Fusion Using Fuzzy Logic, IJSC, Vol.3, No.1, pp. 13-25,February 2012.

[16]Jionghua Teng Suhuan Wang Jingzhou Zhang Xue Wang Coll. of Autom., Northwestern Polytech. Univ.(NPU), Xi'an, China Neuro-fuzzy logic based fusion algorithm of medical images

[17] Mumtaz, A., Masjid, A.Genetic Algorithms and its Applications to Image Fusion. In: IEEE International Conference on Emerging Technologies, Rawalpindi, pp. 6–10 (2008)

[18]Seetha M, MuraliKrishna I.V & Deekshatulu, B.L, (2005) "Data Fusion Performance Analysis Based on Conventional and Wavelet Transform Techniques", IEEE Proceedings on Geoscience and Remote Sensing Symposium, Vol 4, pp. 2842-2845.

[19] X.H. Yang., F.Z.Huang, G.Liu. Urban Remote Image fusion using Fuzzy Rules. International Conference on Machine Learning and Cybernetics, pp. 101-109,2009

[20] V.P.S. Naidu, J.R. Rao. Pixel-level Image Fusion using Wavelets and Principal Component Analysis.Defence Science Journal, pp. 338 -352, 2008.

[21] The Online Resource for Research in Image Fusion, http://www.imagefusion.org